# Apply Distributed CNN on Genomics to accelerate Transcription-Factor TAL1 Motif Prediction

Tasnim Assali
RAMSIS, CRISTAL laboratory
ENSI
Manouba, Tunisia
tasnim.assali@ensi-uma.tn

Zayneb Trabelsi Ayoub
RAMSIS, CRISTAL laboratory
ENSI
Manouba, Tunisia
zayneb.trabelsi@ensi-uma.tn

Sofiane Ouni
RAMSIS, CRISTAL laboratory
ENSI
Manouba, Tunisia
sofiane.ouni@insat.ucar.tn

*Abstract*—Big Data works perfectly along with Deep learning to extract knowledge from a huge amount of data. However, this processing could take a lot of training time. Genomics is a Big Data science with high dimensionality. It relies on deep learning to solve complicated problems in certain diseases like cancer by using different DNA information such as the transcription factor. TAL1 is a transcription factor that is essential for the development of hematopoiesis and of the vascular system. In this paper, we highlight the potential of deep learning in the field of genomics and its challenges such as the training time that takes hours, weeks, and in some cases months. Therefore, we propose to apply a distributed deep learning implementation based on Convolutional Neural Networks (CNN) that showed good results in decreasing the training time and enhancing the accuracy performance with 95% by using multiple GPU and TPU as accelerators. We proved the efficiency of using a distributed strategy based on data-parallelism in predicting the transcription-factor TAL1 motif faster.

## I. Introduction

Genomics is the study of all of a person's genes (the genome), including interactions of those genes with each other and with the person's environment. A genome is an organism's complete set of Deoxyribonucleic acid (DNA). DNA contains information needed to build the entire human body such as the transcription factor binding site (a short nucleotide sequence). It controls when, where, and how efficiently to catalyze the chemical reactions that synthesize RNA (ribonucleic acid), using the gene's DNA as a template. As consequence, it is involved in large number of human diseases such as cancer. TAL1 is a transcription factor, essential for the development of hematopoiesis and of the vascular system. It is one of the most frequently dysregulated oncogenes in T-cell acute lymphoblastic leukemia (T-ALL) with about 30%–50% of patients showing TAL1 overexpression [1].
Besides, researchers admitted that genomics is Big Data science as the work of Stephens et al. [2] demonstrated that genomics is a "four-headed beast"—it is either on par with or the most demanding of the domains analyzed here in terms of data acquisition, storage, distribution, and analysis. This data is characterized by a huge volume and complexity; sequencing a single whole-genome generates more than 100 gigabytes of data. As a consequence, it raises challenges in clinical practice. Therefore, lately researchers start to propose solutions to understand and extract knowledge from this data. They start to manipulate it with machine learning tools that have proven to be very useful in solving bioinformatics-related research questions, especially problems based on classification, clustering, and regression. In particular, deep learning (a subfield of machine learning) shows its efficiency with huge volume and complicated data, but there are several challenges facing these tasks. Various works showed good results but the training time takes days and weeks to get those results and it could influence the performance of the model. On the other hand, distributed machine learning approaches practitioners to shorten model training and inference time by orders of magnitude [3]. Therefore, in this paper, we apply a distributed deep learning implementation that aims to show the efficiency of using distributed deep learning to accelerate the prediction of TAL1 motif in DNA sequences.

The rest of the paper will be organized as follows: Section 2 introduces the background of our work, section 3 is reserved for the related works, section 4 presents our distributed deep learning implementation and section 5 represents the results of experiments.

## II. Background

In this section, we introduce the different fundamental concepts of our work. First, we are going through Deep learning and its concepts; in particular Convolutional neural network. Then we will introduce distributed deep learning: how it works and its challenges.

### A. Deep Learning

Deep learning algorithms have gained huge fame over the last decades due to their capability to solve complicated problems. Researchers in computer science have been experimenting with the neural network since 1950, but two big breakthroughs laid the foundation for nowadays' vast deep learning industry. The first one was in 1986, when Geoffrey Hinttron et al [4], introduced the multilayer Neural Network. To train this network they used the backpropagation algorithm to adjust the output depending on the parameters in each layer but it didn't show its real potential till the second breakthrough that was in 2012, due to the circumstances that lead to collect a huge amount of data helped the researchers to prove the capability of Deep learning such as the work proposed by Alex Krizhevshy and his team, a convolutional neural Network called "AlexNet" that halved the existing error rate on Imagenet visual recognition to 15.3%, [5]. Since then, Deep learning became a famous tool that is capable to handle a large amount of data and solve complicated problems in different fields such as Automated Driving, Medical Research, Aerospace, Defense, etc.

*B. Convolutional Neural Network (CNNs)*

Convolutional neural network (ConvNets or CNNs) is one of the main categories that are well known for working on images and extracting features for classification. CNN was the first variant of multilayer perceptron; it was invented in 1990. They are composed of multiple layers of artificial neurons. Each layer generates several activation functions that are passed on to the next layer. The first layer usually extracts basic features such as horizontal or diagonal edges. This output is passed on to the next layer which detects more complex features such as corners or combinational edges. As we move deeper into the network it can identify even more complex features such as objects, faces, etc. Based on the activation map of the final convolution layer, the classification layer outputs a set of scores (between 0 and 1) that specify the prediction.

*C. Distributed Deep Learning*

Over time, the computing components keep evolving to solve complicated and high dimension issues. Also, Deep learning has become a vital tool to develop new solutions that analyze huge amounts of data such as genomics, which leads to high computational infrastructure solutions such as distributed deep learning. It is used in different fields when we want to speed up our model training process using multiple Graphics processing unit (GPUs). There are mainly two types of distributed training of deep learning: model parallelism and data parallelism as illustrated in the Fig. 1. Model parallelism is used when your model has too many layers to fit into a single GPU and hence different layers are trained on different GPUs. Data parallelism achieves speed-up by splitting the data into different partitions. The model is run simultaneously on multiple GPUs using different data partitions.

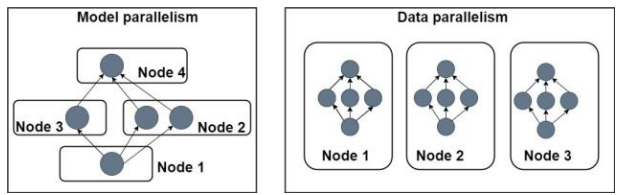

Fig. 1. Distributed Deep learning Models

Data Parallelization in the most used model due to the high cost of Model Parallelization and other reasons [6]. But design it have issues such as the aggregation algorithms that plays a vital role in the distributed learning process. There are two types of aggregation algorithm: the first one is the centralized with a parameter Server and the second one is decentralized such as the algorithm of All-reduce and the algorithm of Gossip. As shown in the Fig. 2, the parameter server stores the model parameters outside of the workers and periodically they report their computed parameters or parameter updates to parameter server(s) (PSs). For All-reduce, there is no parameter server and it mirrors all the model parameters across all workers and workers exchange parameters updates directly via an all-reduce operation. For the algorithm of gossip, there is neither parameter server nor global model. Therefore, every worker communicates updates with their neighbors and the consistency of parameters across all workers is achieved only at the end of the algorithm.

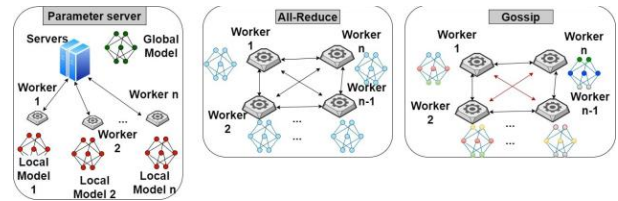

Fig. 2. The aggregation Algorithms

III. RELATED WORKS

In this section, we introduce related works that used deep learning in the genomics fields and works that propose distributed deep learning in order to enhance the performance of deep learning.

*A. Deep learning in the Genomics fields*

There are several works that proposed a solution in genomics fields with deep learning at a different level of genomics as described in Table I. For example, there are works focused on the level of DNA, such as Li et al. [7] who proposed supervised deep learning to identify cis-regulatory regions genome-wide. They identified delineated locations of 300,000 candidate enhancers and 26,000 candidate promoter's genome-wide. As well as Nielsen et al. [8] proposed deep learning to predict the lab-of-origin of a DNA sequence. A CNN was trained on the Add gene plasmid dataset. It achived 77% as accuracy with a 48% validation accuracy and 48% cross-validation accuracy. Also, the training required 21 hours using a single GPU. Other researchers worked on the level of the protein transcription factor (TF), Zhen Cao et al. [9] worked on protein transcription factors and proposed a framework named gkm-DNN to achieve feature representation from high-dimensional gkm-fvs using deep neural networks (DNN). Their experiments show that gkm-DNN made overcomes the drawbacks of high dimensionality, colinearity, and sparsity of gkm-fvs and achieved a better accuracy compared with gkm-SVM in a much shorter training time. Another work on the level of a protein transcription factor by Daniel et al. [10] implemented a convolutional-recurrent neural network model, named FactorNet, to computationally impute the missing binding data. It trains on binding data from reference cell types to make predictions on testing cell types by leveraging a variety of features, including genomics sequences, genome annotations, gene expression, and signal data. Chen et al. [11] proposed a hybrid approach between kernel methods and DNN, convolutional kernel network (CKN) to improve the prediction of TF binding sites. As a result, it outperforms Deep Bind's state-of-the-art convolutional neural networks on a transcription factor binding prediction task while being much faster to train and yielding more stable and interpretable results.

TABLE I
RELATED WORKS: DEEP LEARNING WITH GENOMICS

| Name | Data | Method | Purpose | Results |
|---|---|---|---|---|
| DECRES [7] | ENCODE and FANTOM | FNN | identify cis-regulatory | - |
| gkm-DNN [9] | ChIP-seq | DNN | feature representation | better accuracy then SVM |
| [8] | Addgene plasmid dataset | CNN | predict the lab-of-origin of a DNA sequence. | training accuracy 77%, 48% validation accuracy and 48% cross-validation accuracy |
| FactorNet [10] | ENCODE dataset | CNN | TF prediction. | - |
| [11] | ENCODE ChIP-seq datasets | CKN | prediction of TF | outperforms DeepBind, less sensitive to hyper-parameters |

All those works and other approaches proposed in the literature prove the efficiency of using deep learning in the genomics fields, but there is a critical issue that faces deep learning approaches in genomics. As mentioned by [8], the training required 21 hours using a single NVIDIA GRID K520 GPU and other works need much more time to train the model (for days and months). Deep leaning requires a large amount of data to ensure an efficient solution. If small datasets are employed, over-fitting may occur for a deep learning network with many parameters and as much the volume of the data gets bigger as much the training time get increased. Furthermore, Deep Neural Networks (DNNs) are becoming an important tool in modern computing applications. Accelerating their training is a major challenge and techniques range from distributed algorithms to low-level circuit design. In order to enhance the performance of Deep learning and scaling up the training, the DNN has become a key approach to decrease the training duration and train models in a reasonable time. Thus, various works in literature propose a Parallel and Distributed Deep Learning in different fields.

### B. Parallel and distributed Deep learning

Intel Corporation worked in the field of distributed AI applications and invented BigDL, a framework to support large-scale, distributed training on top of Apache Spark [12]. It offers synchronous data-parallel training to train a deep neural network model across the cluster and aims to achieve better scalability and efficiency in terms of time to quality. Also, it implements an efficient AllReduce-like operation using existing primitives in Spark to support parameter synchronization. The experiment of training Inception-v1 on ImageNet dataset in BigDL made a good result in terms of Computing Performance, Scalability of distributed training, and Efficiency of task scheduling. Another solution has been invented by Uber Technologies Inc, by Alexander Sergeev and Mike Del Balso, named Horovod [13]. It is an open-source library that improves on both obstructions to scaling. It employs efficient inter inter-GPU communication via ring reduction. But there are a few areas that need to be worked on to improve Horovod, including collecting and sharing learnings about adjusting model parameters for distributed deep learning, and Adding examples of very large models. Soulaimane et al. [6], propose Automatic CNN parallelization that fully automates scaling up CNN training to offer distributed deep learning. It is based on RING-ALL reduce parallelism approach and MPI communication protocol. Ring-All is a decentralized synchronous Ring-All reduce data parallelism strategy.

The genomics area demands high computational performance, and deep learning in the last years has shown the potential to satisfy those demands but there are areas that need to be improved. Decreasing the training duration of deep learning throughout scaling up the training process has become one of the most active areas of research making deep learning converge to high-performance computing (HPC) problems. In this paper, we apply a distributed CNN implementation for genomics predictions and shows the efficiency of using distributed deep learning to predict transcription-factor TAL1 motif in DNA sequences. Implement it with different accelerators (GPU and Tensor processing unit (TPU)) and compare the different implementations.

## IV. DISTRIBUTED CNN FOR GENOMICS

Genomics is an interdisciplinary field of biology focusing on the structure, function, evolution, mapping, and editing of genomes. Also, this field contains a massive amount of data that need to be processed to extract knowledge and use it to improve research areas such as cancer. Lately, working with deep learning became trendy with good results but it has limits as the training time. In this paper, we apply a distributed neural network that can discover TAL1 motif in DNA based on learning how to localize a homotypic motif cluster. Our implementation aims to show the efficiency of using distributed deep learning in the genomics field.

### A. The Data

SimDNA is a tool for generating simulated regulatory sequence for use in experiments/analyses. We used it to simulate TAL1 motif density localization in 1500bp (base pairs) long sequence. In total we get 20000 DNA sequences: 10000 positive and 10000 negative sequences (Fig. 3). Our goal is to build a classifier based on convolutional neural networks to distinguish two classes of sequences: positive and negative (depending on the presence or absence of the TAL1 motif). The inputs of our convolutional neural networks are matrix with a one-hot-encoding (Fig. 4). With one-hot-encoding, we convert each categorical value into a new categorical column and assign a binary value of 1 or 0 to those columns. Each integer value is represented as a binary vector. All the values are zero, and the index is marked with a 1. After preparing the data we split it into: 70% for training, 10% for testing, and 20% for validation.

```
'AAACGCCTCAGGATGGTGATTTCACTGCTATTGTCTGC.
AAGATATCTGGCTCCTGATGCAAATCGGCAATAAATAGT.
TGTACGCCGATATATGGACTAAGGACAATCGCAACGCG
ACCTAATTGATTAATCAGCTTCACGTCCAGCTGATTCAG
GGCGTAATCCTCACGTAAAGTGGCAGACCTATAACATAA
```

```
([[[1, 0, 0, 0],
  [0, 0, 1, 0],
  [0, 0, 0, 1],
  [0, 1, 0, 0],
  [0, 0, 1, 0],
```

Fig. 3. DNA sequence

Fig. 4. One-hot encoder

### B. The Distributed Strategy

In this paper, we use a distributed strategy for a deep learning approach by using TensorFlow features. TensorFlow offers to distribute training across multiple GPUs, multiple machines, or TPUs. This feature can help us to distribute our deep learning model and train it with efficiency. For our work, we used one of the TensorFlow strategies called the MirroredStrategy. It offers synchronous training across multiple replicas on one machine. This strategy is typically used for training on one machine with multiple GPUs. For example, a variable created under a Mirrored Strategy is a Mirrored Variable. If no devices are specified in the constructor argument of the strategy, then it will use all the available GPUs. This strategy replicates the variables across all the replicas and keeps them synchronous using an all-reduce algorithm. Also, to implement distributed deep learning we chose data parallelism. Therefore, we define the size of batches for the distributed architecture as follows: buffer size = 10000, shuffle buffer size = 100, Batch size per replica= 64, Batch size = Batch size per replica * number of replicas in synchronous.

### C. The Neural Network Architecture

After choosing the distributing strategy, we identify our Deep learning model. We choose a 2D convolutional neural network (CNN). A CNN learns to recognize patterns that are generally invariant across space, by trying to match the input sequence to several learnable "filters" of a fixed size. In our dataset, the filters will be motifs within the DNA sequences. The CNN may then learn to combine these filters to recognize a larger structure (e.g. the presence or absence of a transcription factor binding site). In our work, the CNN model is defined with the following layers as shown in the Fig. 5. First, we start with the Convolution Layer 2D. We define our convolutional layer to have 15 filters with width = 10, the number of filters is the number of neurons. After the convolution layer, we use the Maximum pooling, it computes the maximum value per-channel in sliding windows of size

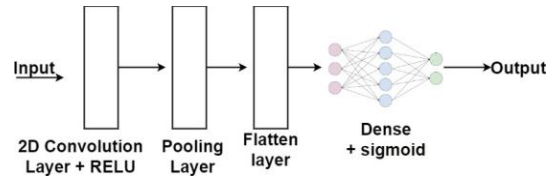

Fig. 5. The CNN model

35. We add the pooling layer because DNA sequences are typically sparse in terms of the number of positions in the sequence that harbor TF motifs. The pooling layer allows us to reduce the size of the output profile of convolutional layers by employing summary statistics. Then we combine the results of the convolution and pooling layers across all 15 filters with the flatten layer and after the output of the fully connected network, we use dense layer with the activation function sigmoid to sculpt the results with a value between 0 and 1. The dense layer is the regular deeply connected neural network layer. Also for compiling our model we used an Adam optimizer that automatically sums gradients across all replicas and use BinaryCrossentropy as a loss function that is used for binary (0 or 1) classification which compute the cross-entropy loss between true labels and predicted labels.

## V. EXPERIMENTS

After preparing our dataset, we start training the model with it. We used Google Colab as our environment, it is a free Jupyter notebook running wholly in the cloud that offers features for distributed strategies with GPU and TPU. A GPU is a performance accelerator and TPU is an artificial intelligence accelerator to accelerate working on the open-source machine learning platform, TensorFlow.

### A. Evaluation Metrics

To measure the performance of our model we calculated the value of:
- Accuracy: the value of correct predictions to the total number of input samples.
- Loss: measure how far an estimated value is from its true value.
- AuROC: measure the model's ability to discriminate between cases
- AuPRC: measure how far the model handles the positive examples.

### B. Training and Evaluation

We started our experiment of distributed CNN with 1 GPU and later implement it with 2 GPUs and 4 GPUs and compare the results. For the first training, we choose the number of epochs = 50, with 1 GPU the model made good results with 93.82% accuracy as illustrated in Fig. 6. Then to improve the training time, we tried the implementation with 2 GPU and 4 GPU and compare it based on the training time. We found that as much the number of GPU increased the training time is considerably decreased as shown in Table II. These results guarantee that working with multiple accelerators such as GPU could decrease the training time of the deep learning model from hours and weeks to minutes and seconds. In literature, several works admit and encourage using TPU with neural networks application, especially with distributed architecture. Therefore, we tried to use it with our distributed work and compared it to the results while using GPU. We prepare the environment to use TPU with Google Colab and then define our distributed strategy with TPU that is defined by 1 worker with 8 cores. For large batch sizes and complex CNN, TPU is the best because of the spatial reuse characteristics of CNNs. As shown with the results (Fig. 7) of our distributed model with TPU, we get better results then the implementation with GPU. The model read the data, split it into batches across all replicas (GPU or TPU). Each replica will carry out training by processing in parallel (the batch size must be divisible by the number of replicas) and after every epoch, the model parameters are passed to aggregation with all-reduce and updates the CNN model and became ready for training in the next epoch. This process will repeat for each epoch until the model reaches convergence when it achieves a state during training in which loss settles to within an error range around the final value. In other words, a model converges when additional training will not improve the model. Therefore, reducing the training time leads to improve the performance as shown with the results of TPU. Hence, the model converged faster and made better accuracy 95.32% as illustrated in Fig. 7, Table II and Table III.

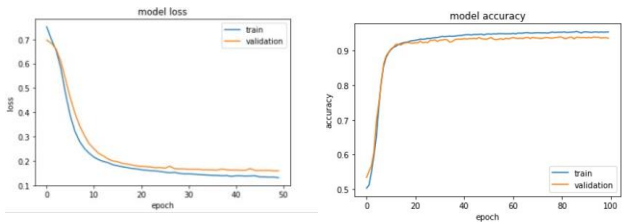

Fig. 6. Training performance with 1 GPU

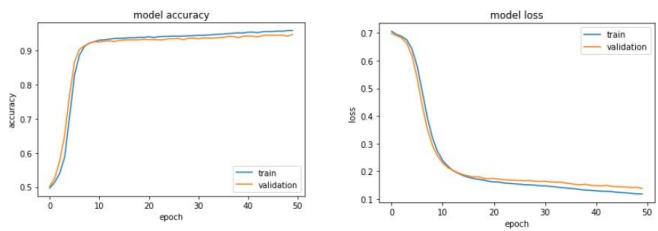

Fig. 7. Training performance with TPU

TABLE II
TRAINING TIME COMPARISON

|  | 1 GPU | 2GPU | 4GPU | TPU |
|---|---|---|---|---|
| Training Time | 133,15s | 99.34s | 58,041s | 56.97s |

## VI. CONCLUSION

In this paper, we highlight the potential of distributed deep learning in the genomics field and its efficiency to accelerate the prediction of transcription factor TAL1 in DNA sequences. Therefore, we implemented a distributed CNN with multiple GPU and TPU with simulated DNA sequences. The experiments showed that we get better results with our distributed strategy based on data-parallelism. Also, we proved that working with multiple GPU made a huge difference in minimizing the training time. Then we made a comparison between working with GPU and TPU and proved that TPU enhances the training time and the performance of the model. In the future, we aim to improve our strategy with larger genomics data and a distributed architecture with a cluster of workers in order to decrease considerably their training time and to extract knowledge from this data faster.

TABLE III
TPU vs GPU

|  | Accuracy | loss | auROC | auPRC |
|---|---|---|---|---|
| 1 GPU | 93,82% | 0.1680 | 98% | 98% |
| 2GPU | 94,54% | 0.1392 | 98% | 98% |
| 4GPU | 95.92% | 0.1036 | 99,1% | 98% |
| TPU | 95.32% | 0.1191 | 98,7% | 98,6% |